\documentclass[conference]{IEEEtran}
\IEEEoverridecommandlockouts
\usepackage{cite}
\usepackage{amsmath,amssymb,amsfonts}
\usepackage{graphicx}
\usepackage{textcomp}
\usepackage{xcolor}

\usepackage{tikz}
\usepackage{pgfplots}
\usepgfplotslibrary{polar}
\pgfplotsset{compat=1.18} 

\definecolor{c_base}{RGB}{200, 200, 200}  
\definecolor{c_truth}{RGB}{70, 130, 180}   
\definecolor{c_ours}{RGB}{205, 92, 92}     

\usepackage{url}            
\usepackage{booktabs}       
\usepackage{microtype}      
\usepackage{multirow}       

\def\BibTeX{{\rm B\kern-.05em{\sc i\kern-.025em b}\kern-.08em
    T\kern-.1667em\lower.7ex\hbox{E}\kern-.125emX}}
\begin{document}

\title{LANCET: Neural Intervention via Structural Entropy for Mitigating Faithfulness Hallucinations in LLMs}

\author{
    \IEEEauthorblockN{
        Chenxu Wang\IEEEauthorrefmark{1}, 
        Chaozhuo Li\IEEEauthorrefmark{1}\IEEEauthorrefmark{3},
        Pengbo Wang\IEEEauthorrefmark{1},
        Litian Zhang\IEEEauthorrefmark{1},
        Songyang Liu\IEEEauthorrefmark{1},
        Ji Qi\IEEEauthorrefmark{1},
        Jiahui Hu\IEEEauthorrefmark{1},\\ 
        Yushan Cai\IEEEauthorrefmark{2}, 
        Hao Zhao\IEEEauthorrefmark{2}
        and Rui Pu\IEEEauthorrefmark{1}
    }
    \IEEEauthorblockA{\IEEEauthorrefmark{1}\textit{Beijing University of Posts and Telecommunications}, Beijing, China}
    \IEEEauthorblockA{\IEEEauthorrefmark{2}\textit{China National Petroleum Corporation}, Beijing, China}
    
    \IEEEauthorblockA{\IEEEauthorrefmark{3}Corresponding author: lichaozhuo@bupt.edu.cn}
    
    \IEEEauthorblockA{Email: chenxu\_w@outlook.com}
}
\maketitle

\begin{abstract}
Large Language Models have revolutionized information processing, yet their reliability is severely compromised by faithfulness hallucinations. While current approaches attempt to mitigate this issue through node-level adjustments or coarse suppression, they often overlook the distributed nature of neural information, leading to imprecise interventions. Recognizing that hallucinations propagate through specific forward transmission pathways like an infection, we aim to surgically block this flow using precise structural analysis. To leverage this, we propose Lancet, a novel framework that achieves precise neural intervention by leveraging structural entropy and hallucination difference ratios. Lancet first locates hallucination-prone neurons via gradient-driven contrastive analysis, then maps their propagation pathways by minimizing structural entropy, and finally implements a hierarchical intervention strategy that preserves general model capabilities. Comprehensive evaluations across hallucination benchmark datasets demonstrate that Lancet significantly outperforms state-of-the-art methods, validating the effectiveness of our surgical approach to neural intervention.
\end{abstract}

\begin{IEEEkeywords}
Large Language Models, Hallucination Mitigation, Neural Intervention, Structural Entropy, Faithfulness Hallucination
\end{IEEEkeywords}

\section{Introduction}

Large Language Models (LLMs) have revolutionized text generation and knowledge retrieval \cite{raffel2020exploring, brown2020language, li2025loki}. However, their deployment in high-stakes domains is impeded by persistent hallucinations—plausible yet unfounded outputs \cite{nguyen2025smoothing, ICLR2025_c98987c5}. This vulnerability poses a critical threat to epistemological integrity, particularly in specialized fields like medical diagnosis \cite{moor2023foundation, salvagno2023artificial}.

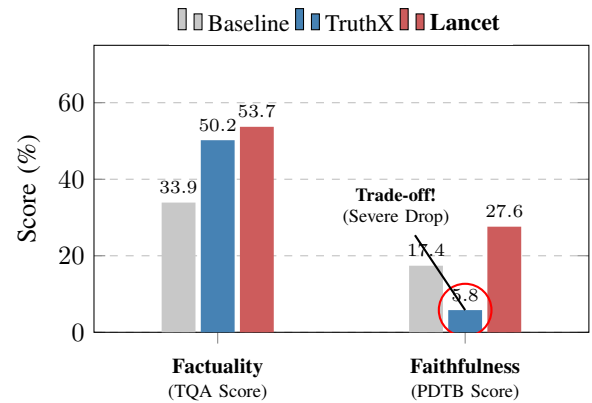
\begin{figure}[t]
    \centering
    \begin{tikzpicture}
        \begin{axis}[
            ybar, 
            bar width=0.45cm,
            width=0.45\textwidth,
            height=5.4cm,
            enlarge x limits=0.5, 
            legend style={
                at={(0.5,1.15)},
                anchor=north,
                legend columns=-1, 
                draw=none, 
                font=\small
            },
            symbolic x coords={Factuality, Faithfulness}, 
            xtick=data,
            xticklabel style={align=center, font=\footnotesize, yshift=-2pt}, 
            xticklabels={
                \textbf{Factuality}\\{\scriptsize (TQA Score)}, 
                \textbf{Faithfulness}\\{\scriptsize (PDTB Score)}
            },
            ylabel={Score (\%)}, 
            nodes near coords, 
            nodes near coords align={vertical},
            every node near coord/.append style={font=\scriptsize},
            ymin=0, ymax=75,
            ymajorgrids=true, 
            grid style=dashed,
        ]

        \definecolor{c_base}{RGB}{200, 200, 200}   
        \definecolor{c_truth}{RGB}{70, 130, 180}   
        \definecolor{c_ours}{RGB}{205, 92, 92}     

        \addplot[fill=c_base, draw=none] coordinates {(Factuality, 33.9) (Faithfulness, 17.4)};
        \addlegendentry{Baseline}

        \addplot[fill=c_truth, draw=none] coordinates {(Factuality, 50.2) (Faithfulness, 5.8)};
        \addlegendentry{TruthX}

        \addplot[fill=c_ours, draw=none] coordinates {(Factuality, 53.7) (Faithfulness, 27.6)};
        \addlegendentry{\textbf{Lancet}}

        \node[draw=red, circle, thick, minimum size=0.7cm] at (axis cs:Faithfulness, 5.8) {};
        
        \node[coordinate, pin={[pin edge={thick, black}, pin distance=1cm, align=center, font=\scriptsize]95:\textbf{Trade-off!}\\(Severe Drop)}] at (axis cs:Faithfulness, 5.8) {};
        
        \end{axis}
    \end{tikzpicture}
    \caption{Performance on LLaMA2-7B-Chat.
    Left: Factuality evaluated on the TruthfulQA benchmark. Right: Faithfulness evaluated on the PDTB benchmark. While TruthX improves factuality at the cost of faithfulness, Lancet excels in both dimensions.}
    \label{fig:motivation_bar}
\end{figure}

While recent efforts prioritize correcting factuality hallucinations—where outputs contradict established world knowledge—they overlook a critical side effect: suppressing these errors often induces faithfulness hallucinations, where the model fails to remain consistent with the provided context \cite{huang2025survey, zhang2023siren}. Our analysis reveals a severe ``Factuality-Faithfulness Trade-off'': interventions designed to enhance factuality often broadly suppress neural activations, which inevitably harms faithfulness \cite{zhang2025explainable}. As illustrated in Fig.~\ref{fig:motivation_bar}, experiments reveal that state-of-the-art methods like TruthX achieve high factuality scores but suffer a catastrophic drop in faithfulness tasks, falling from 17.4 to just 5.8. This confirms that current solutions typically sacrifice context adherence to suppress factual errors, thereby undermining the model's overall utility.

The origin of this trade-off lies in the design limitations of existing mitigation strategies. Predominant approaches, such as TruthX \cite{zhang-etal-2024-truthx} and ITI \cite{10.5555/3666122.3667919}, are typically sample-based, which identify specific neurons based on isolated input samples \cite{qian2024deandeactivatingcoupledneurons, zhang2025activelayercontrastivedecodingreduces}. To correct errors, these methods employ hard mitigation techniques, such as setting activation values to zero or clamping them to a fixed mean \cite{mohammadzadeh2025hallucinationdetoxsensitivitydropout, jiang-etal-2025-neuron}. This strategy assumes that neurons function independently and can be silenced without consequences. However, this approach overlooks a simple fact: neurons in LLMs often do more than one job. The same neuron might help create an error, but also be needed for correct reasoning \cite{zhu2024pollmgraph}. As a result, when these neurons are shut down to fix errors, their useful reasoning abilities are damaged too. This is what causes the severe drop in faithfulness.

However, moving from suppressing entire neurons to a more precise approach presents three major challenges.
First, it is difficult to find the exact source of an error. Because neurons often perform multiple functions, simply looking at their activation values is not enough to determine if they are contributing to an error or to correct reasoning. We need a better way to see how a neuron's behavior changes in different situations \cite{farquhar2024detecting}.
Second, even if the source is found, tracking how the error spreads is also difficult. The connections in the network that spread errors are mixed in with the connections needed for correct reasoning, and there are no clear boundaries. A reliable method is needed to tell these two types of connections apart \cite{chen2024context, li2016structural}.
Third, it is hard to fix the error without causing other problems. A strong correction might damage the model's other abilities, while a weak correction may not stop the error. The main difficulty is designing a method that can adjust the strength of the correction for each neuron. This method must block errors effectively without harming the model's overall faithfulness \cite{ji2023survey}.

To address these challenges, we propose LANCET (Localized Alignment of Neural Connectivity via Entropic Topology), a novel framework designed for precise neural intervention on LLMs. Lancet operates in three stages to systematically isolate and suppress hallucinations. First, it pinpoints source neurons via a gradient-driven contrastive analysis, identifying units with significant activation differentials between factual and hallucinatory contexts. Second, to map the dynamic propagation of misinformation, we introduce the Hallucination Difference Ratio (HDR) as a connectivity metric. By minimizing the Structural Entropy (SE) of the HDR-weighted graph, Lancet partitions the network to structurally separate the connections responsible for errors from those essential for its reasoning abilities. Finally, we implement a hierarchical intervention strategy that scales suppression intensity based on topological proximity to the source. This ensures that hallucinatory behaviors are strictly severed while critical functional pathways remain functionally preserved. Comprehensive evaluations across two benchmarks demonstrate that Lancet significantly outperforms state-of-the-art methods.

Our contributions are:
\begin{enumerate}
    \item We identify the limitations of coarse systemic suppression and propose a topological perspective, treating hallucinations as a structural problem in the network, rather than a defect of individual neurons.

    \item We introduce Lancet, a unified framework that combines gradient-driven contrastive analysis for precise localization with Structural Entropy to structurally identify the pathways that spread misinformation, leaving the rest of the network untouched.

    \item Extensive experiments demonstrate that Lancet effectively resolves the factuality-faithfulness trade-off, significantly outperforming state-of-the-art baselines in mitigating hallucinations while preserving the model's functional integrity.
\end{enumerate}

\section{Theoretical Insight: The Inevitability of Polysemantic Neurons}
\label{sec:theory}

We provide a theoretical argument to demonstrate why node-level interventions are inherently limited in complex models. Let $h \in \mathbb{R}^d$ denote the activation vector at a given layer, and let $g_{\text{fact}} = \nabla_h \mathcal{L}_{\text{fact}}(h)$ and $g_{\text{faith}} = \nabla_h \mathcal{L}_{\text{faith}}(h)$ represent the sensitivity gradients for factuality and faithfulness, respectively. In realistic LLMs, these objectives are intrinsically coupled; improving knowledge retrieval often influences the reasoning logic required to present that knowledge. We capture this interdependence by assuming the gradients are generically \textit{non-orthogonal} over the data distribution $\mathcal{D}$:
\begin{equation}
    \mathbb{E}_{h \sim \mathcal{D}} \left[ \langle g_{\text{fact}}(h), g_{\text{faith}}(h) \rangle \right] \neq 0.
    \label{eq:correlation}
\end{equation}

To understand the structural implications of this correlation, consider the hypothesis that neuron abilities are \textit{strictly independent}. This ideal scenario would imply that the set of all neuron indices $\mathcal{I} = \{1,\dots,d\}$ can be partitioned into two disjoint subsets, $\mathcal{I}_{\text{fact}}$ and $\mathcal{I}_{\text{faith}}$, such that each neuron acts as a dedicated specialist for at most one task. Mathematically, this enforces a strict exclusion condition: for any neuron $i$, if it is active for factuality, it must be silent for faithfulness. This can be expressed as the vanishing component-wise product for all neurons:
\begin{equation}
    \forall i \in \mathcal{I}, \quad g_{\text{fact},i}(h) \cdot g_{\text{faith},i}(h) = 0.
\end{equation}

Under this hypothesis, we can analyze the inner product of the gradients by decomposing the summation over the partitioned indices. The total inner product would split into two orthogonal terms:
\begin{equation}
\begin{aligned}
    \langle g_{\text{fact}}, g_{\text{faith}} \rangle &= \sum_{i \in \mathcal{I}_{\text{fact}}} g_{\text{fact},i} \cdot g_{\text{faith},i} + \sum_{j \in \mathcal{I}_{\text{faith}}} g_{\text{fact},j} \cdot g_{\text{faith},j} \\
    &= \sum_{i \in \mathcal{I}_{\text{fact}}} (g_{\text{fact},i} \cdot 0) + \sum_{j \in \mathcal{I}_{\text{faith}}} (0 \cdot g_{\text{faith},j}) \\
    &= 0.
\end{aligned}
\end{equation}
Taking the expectation yields $\mathbb{E}[\langle g_{\text{fact}}, g_{\text{faith}} \rangle] = 0$, which directly contradicts the generic non-orthogonality condition established in Eq.~\eqref{eq:correlation}. This contradiction proves that the intersection of active neurons must be non-empty, mathematically necessitating the existence of polysemantic neurons.

This result allows us to quantify the risk of collateral damage inherent in coarse interventions. Consider a standard node-level intervention on a polysemantic neuron $k$ intended to suppress a factual error (modifying $h_k$ by $\Delta h_k$). The unintended impact on the faithfulness loss can be approximated by the first-order Taylor expansion:
\begin{equation}
    \Delta \mathcal{L}_{\text{faith}} \approx g_{\text{faith},k} \cdot \Delta h_k \neq 0.
\end{equation}
Since the sensitivity $g_{\text{faith},k}$ is non-zero for these shared neurons, any modification intended for factuality inevitably manifests as a perturbation in $\mathcal{L}_{\text{faith}}$. This theoretical bound explains the trade-off observed empirically and necessitates the shift to the topology-aware intervention proposed in Lancet, which targets pathways rather than individual nodes.

\section{Methodology: The Lancet Framework}
\label{sec:method}
\begin{figure*}[t] 
    \centering
    \includegraphics[width=0.85\textwidth]{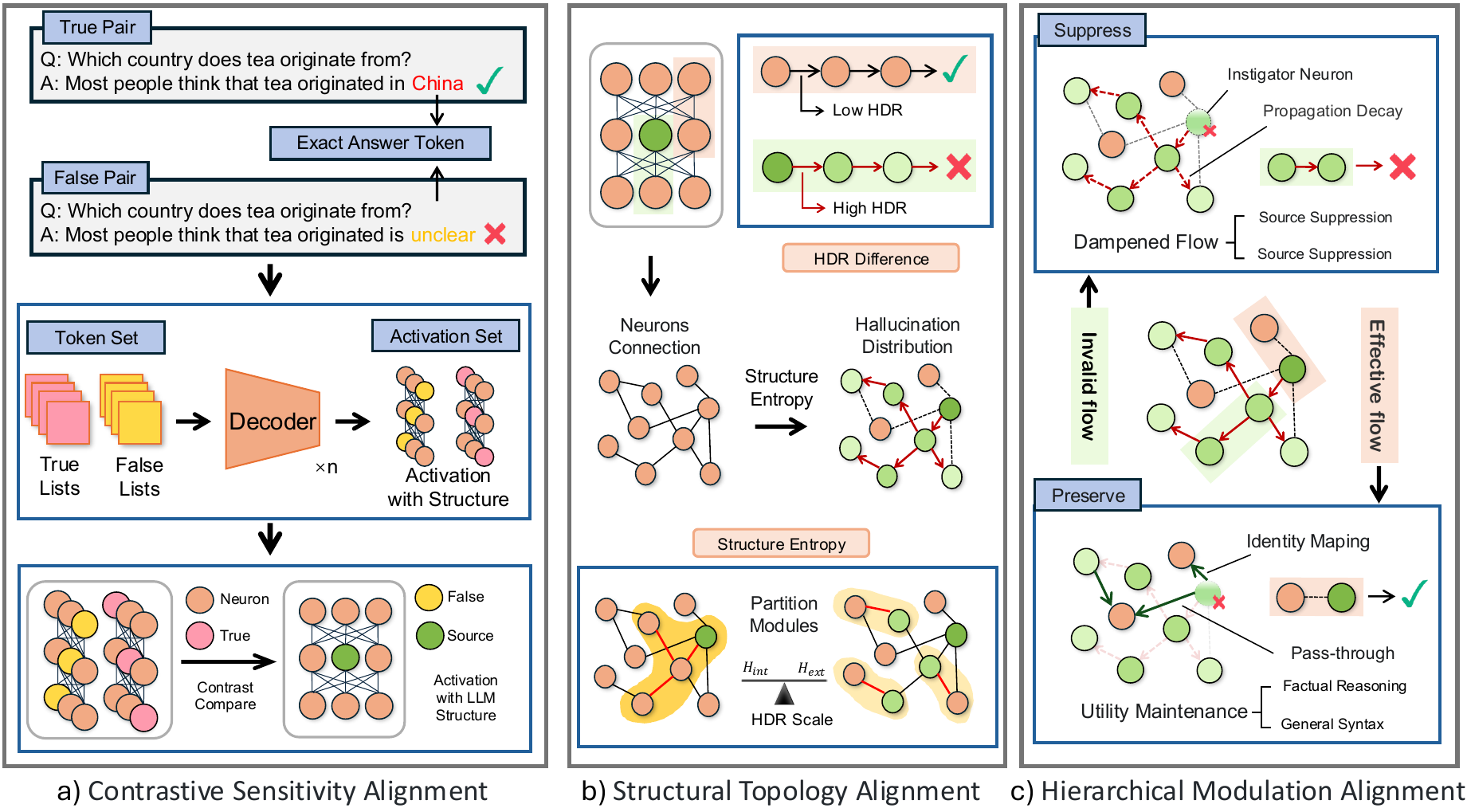} 
    \caption{Overview of the lancet Framework. Lancet orchestrates a three-stage intervention: (1) \textbf{Contrastive Sensitivity Alignment} locates instigator neurons by analyzing gradient differentials; (2) \textbf{Structural Topology Alignment} maps propagation pathways via Structural Entropy minimization; (3) \textbf{Hierarchical Modulation Alignment} executes precise surgical suppression based on topological proximity.}
    \label{fig:lancet}
\end{figure*}

We propose Lancet, a unified framework designed to surgically mitigate faithfulness hallucinations by topologically isolating misinformation flows from general reasoning circuits. Unlike traditional methods that treat neurons as independent units, Lancet adopts a systemic perspective, viewing hallucinations as propagative phenomena. As systematically illustrated in Fig.~\ref{fig:lancet}, the framework operates through three strictly aligned phases: 
(1) Pinpointing the genesis of errors via contrastive sensitivity alignment;
(2) Mapping the dynamic propagation of errors by aligning topological boundaries with functional flows;
(3) Executing targeted corrections through a hierarchical modulation strategy scaled by topological proximity.

\subsection{Contrastive Sensitivity Alignment}
To precisely identify the origin of hallucinations, we employ a Contrastive Sensitivity Alignment strategy. This approach is predicated on the hypothesis that specific ``instigator'' neurons exhibit divergent sensitivity patterns when the model transitions from faithful factual generation to hallucinatory fabrication. Conventional methods often rely on static activation magnitudes to select neurons; however, such metrics fail to capture the functional causal link between a neuron and the model's output errors. To address this, we align our selection criteria with the dynamic gradient differentials observed between distinct contexts.

We quantify the functional influence of each neuron by computing its gradient-based importance score. This metric serves as a proxy for the neuron's contribution to the final decision logic. For a given dataset $D$, the sensitivity of the model's output to a specific neuron $u$ is formally defined as the expected magnitude of the parameter-loss gradient product:
\begin{equation}
    I_{u}^D = \mathbb{E}_{s \sim D} \left| \theta_{u} \cdot \nabla_{\theta_{u}} \mathcal{L}(s) \right|
\end{equation}
where $\theta_{u}$ represents the weight parameters of neuron $u$, and $\mathcal{L}(s)$ denotes the standard loss function evaluated on sample $s$.

To capture the behavioral shift associated with hallucinations, we systematically evaluate this metric across two contrasting distributions: the hallucinatory response set ($I_{u}^{\text{hall}}$) and the ground-truth factual set ($I_{u}^{\text{fact}}$). The divergence in the neuron's functional role is then encapsulated by the sensitivity difference:
\begin{equation}
    \Delta I_{u} = I_{u}^{\text{hall}} - I_{u}^{\text{fact}}
\end{equation}
A significant positive value of $\Delta I_{u}$ indicates that neuron $u$ becomes disproportionately active and influential specifically when the model is driving misinformation. We rank all candidate neurons within the layer based on this alignment score and designate the top fraction as the ``instigator set'' $\mathcal{N}_{src}$. Crucially, to preserve the model’s general utility and avoid damaging general cognitive circuitry, we explicitly filter out neurons that demonstrate high baseline sensitivity on general reasoning benchmarks like PDTB. While this sensitivity is calculated using the same gradient-driven metric, it is evaluated on a general task distribution rather than the misinformation contrast, thereby ensuring the specificity of our target set.

\subsection{Structural Topology Alignment}
Identifying the root instigators is a necessary but insufficient step, as misinformation propagates through complex, entangled downstream pathways. To address this challenge, we generate a structured view of the network that aligns topological boundaries with functional propagation flows. We achieve this by constructing a weighted directed graph $G=(\mathcal{N}_{src}, E, \mathcal{A})$ where $E$ represents functional pathways with weights $\mathcal{A}$ reflecting the intensity of misinformation flow and dynamical complexity, and partitioning it using the Structural Entropy framework.

\paragraph{HDR-Based Graph Construction.}
To enforce topological alignment, the edge weights in Lancet must reflect the \textit{tendency} to propagate hallucinations rather than generic connectivity. We introduce the Hallucination Difference Ratio (HDR) as the affinity weight $a_{uv}$ between neurons $u$ and $v$. This metric quantifies the amplification of correlation during hallucinatory states:
\begin{equation}
    a_{uv} = \text{HDR}_{uv} = \frac{\left| \rho_{uv}^{hall} - \rho_{uv}^{fact} \right|}{\max(|\rho_{uv}^{fact}|, \epsilon)}
\end{equation}
where $\rho_{uv}^{D}$ denotes the Pearson correlation coefficient of activations on dataset $D$, and $\epsilon$ is a small constant for numerical stability. A high HDR value signifies that the connection between $u$ and $v$ becomes anomalously active or synchronized during hallucinations, effectively serving as a primary conduit for misinformation flow.

\paragraph{Minimizing Structural Entropy.}
With the graph constructed, our goal is to delineate functional modules that isolate these high-HDR pathways. We employ the minimization of Structural Entropy \cite{li2016structural} as our optimization objective. Let $G = (\mathcal{N}_{src},E,\mathcal{A})$ be the HDR-weighted activation graph and $\mathcal{P} = \{X_1,\dots,X_m\}$ a partition of $V$ into disjoint modules. We define $\mathrm{vol}(G) = \sum_{v_i \in \mathcal{N}_{src}} d_i$ as the total volume of the graph, where $d_i$ is the weighted degree of node $v_i$. Similarly, $\mathrm{vol}(X)$ represents the volume of module $X$, and $\mathrm{cut}(X)$ denotes the total weight of edges bridging $X$ and the rest of the graph.

The structural entropy $\mathcal{E}^{\mathcal{P}}(G)$ is defined as the summation of the internal entropy ($H_{\text{int}}$) and external entropy ($H_{\text{ext}}$) across all modules:
\begin{equation}
    \mathcal{E}^{\mathcal{P}}(G)
    = \sum_{X \in \mathcal{P}}
      \bigl( H_{\text{int}}(X) + H_{\text{ext}}(X) \bigr).
    \label{eq:se-main}
\end{equation}
These components are mathematically formulated as:
\begin{align}
    H_{\text{int}}(X)
    &= - \sum_{v_i \in X}
       \frac{d_i}{\mathrm{vol}(G)}
       \log_2 \frac{d_i}{\mathrm{vol}(X)}, \label{eq:se-int} \\
    H_{\text{ext}}(X)
    &= - \frac{\mathrm{cut}(X)}{\mathrm{vol}(G)}
       \log_2 \frac{\mathrm{vol}(X)}{\mathrm{vol}(G)}. \label{eq:se-ext}
\end{align}
Physically, $H_{\text{int}}$ captures the stability of random walks within a module, while $H_{\text{ext}}$ measures the uncertainty associated with inter-module transitions. Minimizing $\mathcal{E}^{\mathcal{P}}(G)$ yields a partition where nodes within modules are tightly coupled by high-HDR edges, while connections between modules are sparse.

We seek an approximately optimal partition $\mathcal{P}^*$ via a two-stage iterative algorithm: (1) \textbf{Merging}, which recursively combines neuron modules that yield the largest reduction in structural entropy; and (2) \textbf{Refinement}, which fine-tunes the boundaries by reassigning individual neurons to maximize the net entropy gain. The resulting partition $\mathcal{P}^*$ topologically separates the ``infected'' propagation zones from the healthy reasoning network.

\subsection{Hierarchical Modulation Alignment}
Based on the derived topology $\mathcal{P}^*$, we implement a Hierarchical Modulation Alignment strategy to surgically intervene in the network. Unlike uniform suppression which risks damaging model capabilities, this mechanism calibrates the suppression intensity for each neuron, aligning the intervention force with the neuron's topological role in the misinformation propagation chain.

We formulate the intervention by establishing a topological suppression factor $\alpha_{u}$ for each neuron. This factor integrates the local severity of incoming flows with a spatial decay relative to the source, ensuring a smooth gradient of intervention:
\begin{equation}
    \alpha_{u} = \alpha_0 \max_{v \to u} \text{HDR}_{vu} \, e^{-\lambda d_u}
\end{equation}
where $\alpha_0$ represents the baseline intensity and $d_u$ denotes the geodesic distance from neuron $u$ to the instigator set $\mathcal{N}_{\text{src}}$ within the graph.

Functionally, the maximization term $\max_{v \to u} \text{HDR}_{vu}$ acts as a worst-case estimator for the influx of misinformation, identifying the primary conduit through which hallucinations propagate to neuron $u$. Concurrently, the exponential term $e^{-\lambda d_u}$ establishes a topological quarantine zone controlled by the decay rate $\lambda$. This design ensures that the suppression force remains spatially localized around the instigators, rapidly attenuating as it reaches general-purpose regions of the network.

The intervention is finally executed as a topology-aware parameter rescaling:
\begin{equation}
    \tilde{\theta}_{u} = (1 - \alpha_{u}) \theta_{u}
\end{equation}
To ensure the model retains its general capabilities, we enforce specific boundary rules. For the root instigator neurons $\mathcal{N}_{\text{src}}$, the suppression factor $\alpha_{u}$ is set to one. This strictly cuts off the source of the hallucination. Conversely, for neurons belonging to critical reasoning pathways $\mathcal{C}_{\text{crit}}$, the suppression factor is fixed at zero. We explicitly define these critical pathways as the set of neurons with the highest importance scores on the factual dataset, which corresponds to the top ranked units derived in the first stage. By protecting these high-value units, our method creates a smooth adjustment across the network. It transitions from hard suppression at the source to soft dampening in the surrounding areas, while leaving essential functions completely unchanged.

\section{Experiments and Results}

\begin{table*}[t]
    \caption{Cross-model comparison and Lancet performance on PDTB and TruthfulQA benchmarks.}
    \label{tab:main-results}
    \centering

    \renewcommand{\arraystretch}{1.05}   
    \setlength{\tabcolsep}{4pt}          

    \newcommand{\modelLLaMA}{%
      \rule{0pt}{11.2ex}\hspace{1em}\rotatebox[origin=c]{90}{LLaMA2-7B-Chat}}
    \newcommand{\modelDPSK}{%
      \rule{0pt}{8.5ex}\hspace{1em}\rotatebox[origin=c]{90}{DeepSeek-llm-7B}}

    {\small
    \begin{tabular*}{\textwidth}{@{\hspace{0.6em}\extracolsep{\fill}}%
                                 l l@{\hspace{0.8em}}|cccc|ccccc}
        \toprule[1.2pt]
        \multirow{2}{*}{\textbf{Method}} & \multirow{2}{*}{\textbf{Model}} &
        \multicolumn{4}{c}{\textbf{PDTB}} &
        \multicolumn{5}{c}{\textbf{TruthfulQA}} \\
        \cmidrule(r){3-6} \cmidrule(r){7-11}
        & & \textbf{Overall} & \textbf{Targeted} &
            \textbf{Counterfactual} & \textbf{Consistency} &
            \textbf{True*Info} & \textbf{MC1} & \textbf{MC2} &
            \textbf{True} & \textbf{Info} \\
        \midrule
        Baseline
          & \multirow{9}{*}{\modelLLaMA}
          & 17.4 & 79.3 & 27.1 & 81.2 & 57.6 & 33.9 & 51.3 & 66.9 & 86.1 \\
        \midrule
        FT-PDTB           & & 23.7 & 81.5 & 34.9 & 83.2 & 46.1 & 30.7 & 45.1 & 62.8 & 73.3 \\
        FT-TruthfulQA     & & 12.7 & 69.2 & 22.7 & 80.6 & 62.5 & 49.7 & 63.2 & 69.0 & 90.6 \\
        CAD               & & 20.0 & 70.5 & 35.4 & 80.1 & 54.2 & 22.0 & 50.8 & 69.0 & 78.5 \\
        DoLa              & & 15.9 & 78.0 & 25.1 & 81.2 & 59.2 & 33.3 & 60.9 & 67.6 & 87.5 \\
        ITI               & & 14.8 & 64.5 & 28.7 & 79.8 & 59.9 & 33.9 & 52.0 & 68.9 & 87.0 \\
        TruthX            & & 5.8  & 75.6 & 9.6  & 80.2 & 63.6 & 50.2 & 70.5 & 70.8 & 89.7 \\
        \textbf{Lancet (Ours)} & & \textbf{27.6} & \textbf{81.9} & \textbf{37.4} & \textbf{90.4} & \textbf{67.8} & \textbf{53.7} & \textbf{72.8} & \textbf{71.9} & \textbf{94.3} \\
        \bottomrule[1pt]
    \end{tabular*}

    \vspace{1.5mm}

    \begin{tabular*}{\textwidth}{@{\hspace{0.6em}\extracolsep{\fill}}%
     l l@{\hspace{0.8em}}|cccc|ccccc}
        \toprule
        \multirow{2}{*}{\textbf{Method}} & \multirow{2}{*}{\textbf{Model}} &
        \multicolumn{4}{c}{\textbf{PDTB}} &
        \multicolumn{5}{c}{\textbf{TruthfulQA}} \\
        \cmidrule(r){3-6} \cmidrule(r){7-11}
        & & \textbf{Overall} & \textbf{Targeted} &
            \textbf{Counterfactual} & \textbf{Consistency} &
            \textbf{True*Info} & \textbf{MC1} & \textbf{MC2} &
            \textbf{True} & \textbf{Info} \\
        \midrule
        Baseline
          & \multirow{9}{*}{\modelDPSK}
          & 15.1 & 21.2 & 84.0 & 85.1 & 59.5 & 37.9 & 55.7 & 70.1 & 84.8 \\
        \midrule
        FT-PDTB           & & 20.1 & 27.8 & 86.2 & 83.7 & 46.8 & 33.1 & 46.2 & 65.1 & 71.8 \\
        FT-TruthfulQA     & & 3.5  & 5.0  & 82.4 & 83.9 & 64.3 & 49.4 & 67.9 & 73.3 & 87.8 \\
        CAD               & & 15.6 & 23.4 & 80.9 & 82.4 & 49.3 & 33.3 & 54.9 & 70.0 & 70.5 \\
        DoLa              & & 10.1 & 14.6 & 83.1 & 83.4 & 60.1 & 29.9 & 7.5  & 70.3 & 85.5 \\
        ITI               & & 15.7 & 22.2 & 83.5 & 84.9 & 62.9 & 38.6 & 57.9 & 71.3 & 88.2 \\
        \textbf{Lancet (Ours)} & & \textbf{23.7} & \textbf{30.1} & \textbf{89.1} & \textbf{88.2} & \textbf{72.8} & \textbf{49.8} & \textbf{68.6} & \textbf{79.5} & \textbf{91.6} \\
        \bottomrule[1pt]
    \end{tabular*}
    }

    \vspace{-1mm}
\end{table*}

\subsection{Datasets}
We validate Lancet on two benchmarks. PDTB \cite{webber2019penn} assesses discourse faithfulness via the DISQ metric \cite{miao-etal-2024-discursive}, which aggregates Targeted accuracy, Counterfactual robustness, and logical Consistency into a unified Overall score. TruthfulQA \cite{lin-etal-2022-truthfulqa} evaluates factuality through multiple-choice accuracy (MC1/MC2) and generative quality (Truthfulness and Informativeness), synthesized into the composite True*Info metric.

\subsection{Baselines}
We benchmark Lancet against representative methods spanning three paradigms: Supervised Fine-Tuning on target datasets to establish performance upper bounds; decoding strategies including CAD~\cite{shi-etal-2024-trusting} and DoLa~\cite{chuang2023dola} that manipulate output logits; and state-of-the-art inference-time intervention techniques such as ITI~\cite{10.5555/3666122.3667919}, and TruthX~\cite{zhang-etal-2024-truthx}.

\subsection{Implementation Details}
Experiments are conducted on LLaMA2-7B-Chat~\cite{touvron2023llama} and DeepSeek-llm-7B \cite{bi2024deepseek} loaded in half-precision. For lancet, the instigator selection ratio is optimized on a hold-out validation set, while Structural Entropy minimization employs the standard partitioning algorithm. All baseline results are reproduced using their official implementations under identical experimental settings to ensure fair comparison.

\subsection{Main Results}

Table \ref{tab:main-results} confirms that baselines like TruthX suffer a severe factuality-faithfulness trade-off, sacrificing reasoning for factuality. In contrast, Lancet achieves simultaneous improvements across both dimensions. This superiority stems from Structural Entropy, which isolates misinformation flow more precisely than subspace projections. Moreover, Lancet demonstrates robust generalization on DeepSeek, maintaining stability where other strategies fail.

\subsection{Ablation Study}

Table \ref{tab:ablation} validates the contribution of Lancet's components across LLaMA2 and DeepSeek. Removing the Hallucination Difference Ratio causes the sharpest drop in faithfulness, confirming that activation magnitude alone is insufficient for precise identification. Similarly, excluding Structural Entropy notably degrades reasoning, particularly on DeepSeek, validating that topological mapping is indispensable for blocking misinformation propagation. Furthermore, replacing hierarchical modulation with uniform suppression yields suboptimal results, demonstrating that aligning intervention intensity with topological severity is critical for balancing mitigation with general capabilities.

\begin{table}[t]
    \caption{Ablation study of Lancet components on LLaMA2 and DeepSeek.}
    \label{tab:ablation}
    \centering
    \footnotesize
    \renewcommand{\arraystretch}{1.1}
    \begin{tabular*}{\columnwidth}{@{\extracolsep{\fill}}l c c c c}
        \toprule
        \multirow{2}{*}{\textbf{Variant}} & 
        \multicolumn{2}{c}{\textbf{LLaMA2-7B}} & 
        \multicolumn{2}{c}{\textbf{DeepSeek-7B}} \\
        \cmidrule(r){2-3} \cmidrule(l){4-5}
         & \textbf{PDTB} & \textbf{TQA} & \textbf{PDTB} & \textbf{TQA} \\
        \midrule
        Baseline          & 17.4 & 57.6 & 15.1 & 59.5 \\
        \midrule
        w/o HDR           & 22.8 & 63.1 & 20.0 & 64.0 \\
        w/o SE            & 21.7 & 64.8 & 18.5 & 69.0 \\
        w/o Hier.         & 22.4 & 64.4 & 21.5 & 68.5 \\
        \textbf{Full Lancet}       & \textbf{27.6} & \textbf{67.8} & \textbf{23.7} & \textbf{72.8} \\
        \bottomrule
    \end{tabular*}
\end{table}

\subsection{Parameter Sensitivity Analysis}

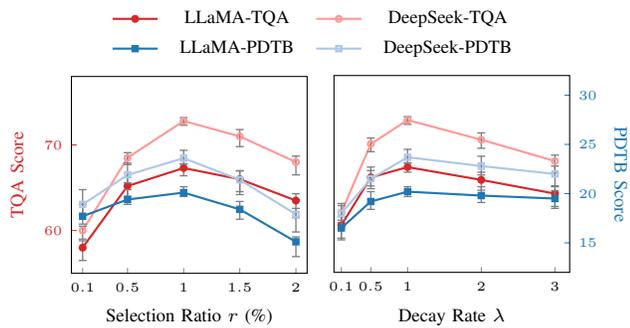
\begin{figure}[t]
    \centering
    \definecolor{tqaLLaMA}{HTML}{D62728}   
    \definecolor{tqaDS}{HTML}{FF9896}      
    \definecolor{pdtbLLaMA}{HTML}{1F77B4}  
    \definecolor{pdtbDS}{HTML}{AEC7E8}     
    \definecolor{lancetGray}{HTML}{7F7F7F} 

    \pgfplotsset{
        myplotstyle/.style={
            width=0.26\textwidth,
            height=4.2cm,
            xlabel near ticks,
            font=\scriptsize,
            tick label style={font=\tiny},
            axis lines=box,         
            grid=none,
            tick align=inside,
            major tick length=2pt,
            error bars/y dir=both,
            error bars/y explicit,
            error bars/error bar style={solid, line width=0.5pt, lancetGray},
            error bars/error mark options={rotate=90, mark size=1.5pt, line width=0.5pt}
        }
    }

    \pgfplotslegendfromname{dual_legend}
    \vspace{0.1cm}

    \begin{tikzpicture}
        \begin{axis}[
            myplotstyle,
            xtick pos=bottom,
            ytick pos=left,
            yticklabel pos=left,
            ylabel={TQA Score},
            y label style={at={(axis description cs:-0.15,.5)},anchor=south, color=tqaLLaMA},
            ytick style={color=tqaLLaMA},
            yticklabel style={color=tqaLLaMA, font=\tiny},
            xlabel={Selection Ratio $r$ (\%)},
            xtick={0.1, 0.5, 1.0, 1.5, 2.0},
            xmin=0, xmax=2.1,
            ymin=55, ymax=78,
            legend columns=2,
            legend to name=dual_legend,
            legend style={
                draw=none,
                fill=none,
                font=\scriptsize,
                column sep=6pt,
                row sep=1pt
            }
        ]

        \addplot[mark=*, mark size=1.1pt, color=tqaLLaMA, thick] coordinates {
            (0.1, 58.0) -= (0, 1.5) += (0, 0.8)
            (0.5, 65.2) -= (0, 1.2) += (0, 1.0)
            (1.0, 67.3) -= (0, 0.9) += (0, 0.5)
            (1.5, 66.0) -= (0, 1.8) += (0, 1.0)
            (2.0, 63.5) -= (0, 2.0) += (0, 0.8)
        };
        \addlegendentry{LLaMA-TQA}

        \addplot[mark=o, mark size=1.1pt, color=tqaDS, thick] coordinates {
            (0.1, 60.0) -= (0, 1.0) += (0, 0.5)
            (0.5, 68.5) -= (0, 0.8) += (0, 0.6)
            (1.0, 72.8) -= (0, 0.5) += (0, 0.4)
            (1.5, 71.0) -= (0, 1.2) += (0, 0.8)
            (2.0, 68.0) -= (0, 1.5) += (0, 0.7)
        };
        \addlegendentry{DeepSeek-TQA}

        \addlegendimage{mark=square*, mark size=1.0pt, color=pdtbLLaMA, thick}
        \addlegendentry{LLaMA-PDTB}
        \addlegendimage{mark=square, mark size=1.0pt, color=pdtbDS, thick}
        \addlegendentry{DeepSeek-PDTB}

        \end{axis}

        \begin{axis}[
            myplotstyle,
            axis lines=none,
            xmin=0, xmax=2.1,
            ymin=12, ymax=32,
            xtick=\empty,
            ytick pos=right,
            yticklabel pos=right,
            yticklabel=\empty,
            ytick style={color=pdtbLLaMA},
        ]
        \addplot[mark=square*, mark size=1.0pt, color=pdtbLLaMA, thick] coordinates {
            (0.1, 17.8) -= (0, 0.8) += (0, 1.2)
            (0.5, 19.5) -= (0, 0.5) += (0, 1.0)
            (1.0, 20.2) -= (0, 0.4) += (0, 0.6)
            (1.5, 18.5) -= (0, 1.0) += (0, 0.8)
            (2.0, 15.2) -= (0, 1.5) += (0, 0.5)
        };
        \addplot[mark=square, mark size=1.0pt, color=pdtbDS, thick] coordinates {
            (0.1, 19.0) -= (0, 1.0) += (0, 1.5)
            (0.5, 22.0) -= (0, 0.8) += (0, 1.2)
            (1.0, 23.7) -= (0, 0.6) += (0, 0.8)
            (1.5, 21.5) -= (0, 1.2) += (0, 0.9)
            (2.0, 18.0) -= (0, 1.8) += (0, 0.6)
        };
        \end{axis}
    \end{tikzpicture}%
    \hspace{0.15cm}%
    \begin{tikzpicture}
        \begin{axis}[
            myplotstyle,
            axis lines=none,
            xmin=0, xmax=3.2,
            ymin=55, ymax=78,
            xtick=\empty,
            ytick pos=left,
            yticklabel pos=left,
            yticklabel=\empty,
            ytick style={color=tqaLLaMA},
        ]
        \addplot[mark=*, mark size=1.1pt, color=tqaLLaMA, thick] coordinates {
            (0.1, 60.5) -= (0, 1.5) += (0, 1.0)
            (0.5, 66.1) -= (0, 1.0) += (0, 0.8)
            (1.0, 67.3) -= (0, 0.6) += (0, 0.5)
            (2.0, 65.8) -= (0, 1.2) += (0, 0.9)
            (3.0, 64.2) -= (0, 1.5) += (0, 0.8)
        };
        \addplot[mark=o, mark size=1.1pt, color=tqaDS, thick] coordinates {
            (0.1, 62.0) -= (0, 1.2) += (0, 0.8)
            (0.5, 70.0) -= (0, 0.9) += (0, 0.7)
            (1.0, 72.8) -= (0, 0.5) += (0, 0.4)
            (2.0, 70.5) -= (0, 1.0) += (0, 0.8)
            (3.0, 68.0) -= (0, 1.4) += (0, 0.7)
        };
        \end{axis}

        \begin{axis}[
            myplotstyle,
            xtick pos=bottom,
            ytick pos=right,
            yticklabel pos=right,
            xlabel={Decay Rate $\lambda$},
            ylabel={PDTB Score},
            y label style={at={(axis description cs:1.15,.5)},anchor=south, rotate=180, color=pdtbLLaMA},
            ytick style={color=pdtbLLaMA},
            yticklabel style={color=pdtbLLaMA, font=\tiny},
            xtick={0.1, 0.5, 1.0, 2.0, 3.0},
            xmin=0, xmax=3.2,
            ymin=12, ymax=32
        ]
        \addplot[mark=square*, mark size=1.0pt, color=pdtbLLaMA, thick] coordinates {
            (0.1, 16.5) -= (0, 1.2) += (0, 0.8)
            (0.5, 19.2) -= (0, 0.8) += (0, 1.0)
            (1.0, 20.2) -= (0, 0.5) += (0, 0.5)
            (2.0, 19.8) -= (0, 0.7) += (0, 0.9)
            (3.0, 19.5) -= (0, 1.0) += (0, 0.8)
        };
        \addplot[mark=square, mark size=1.0pt, color=pdtbDS, thick] coordinates {
            (0.1, 18.0) -= (0, 1.5) += (0, 1.0)
            (0.5, 21.5) -= (0, 1.0) += (0, 1.2)
            (1.0, 23.7) -= (0, 0.6) += (0, 0.8)
            (2.0, 22.8) -= (0, 0.9) += (0, 1.0)
            (3.0, 22.0) -= (0, 1.2) += (0, 0.8)
        };
        \end{axis}
    \end{tikzpicture}

    \caption{Cross-model parameter sensitivity analysis for LLaMA2 and DeepSeek on TruthfulQA (red tones) and PDTB (blue tones). Error bars are asymmetric standard deviations.}
    \label{fig:sensitivity}
\end{figure}

We investigate the robustness of Lancet regarding the instigator selection ratio $r$ and spatial decay factor $\lambda$, as detailed in Fig. \ref{fig:sensitivity}. The trajectory for the selection ratio reveals a delicate balance: while expanding the surgical scope initially enhances truthfulness, excessive selection encroaches upon general reasoning circuitry, leading to a decline in faithfulness metrics. This confirms that the optimal ratio represents the precise boundary of the infected neuron set. Parallel behavior is observed for the spatial decay factor, where the performance peak validates the existence of a specific topological radius for effective intervention. Deviations toward either overly diffused or strictly localized suppression diminish the overall synergy, demonstrating that the intervention intensity must be geometrically matched to the natural propagation of misinformation.

\section{Conclusion}

In this work, we presented Lancet, a topological intervention framework that addresses faithfulness hallucinations by modeling them as directed ``infectious'' flows. By leveraging Hallucination Difference Ratio and Structural Entropy, Lancet shifts the paradigm from coarse suppression to precise, surgery-like modulation, effectively excising misinformation pathways. Our empirical results verify that this approach resolves the critical trade-off between hallucination mitigation and general reasoning preservation, achieving state-of-the-art performance. While current reliance on paired data and graph partitioning overhead presents scalability challenges, future integration of unsupervised instigator detection and lightweight approximations aims to further streamline the process. Ultimately, Lancet establishes a rigorous geometric foundation for neural intervention, offering both a theoretical lens for decoding information flow and a practical pathway toward reliable and functionally robust LLMs.

\bibliographystyle{IEEEbib}
\bibliography{references}

\end{document}